\documentclass[acmtog]{acmart}

\usepackage{graphicx}
\usepackage{tikz}
\usepackage{comment}
\usepackage{amsmath} 
\usepackage{makecell}

\AtBeginDocument{%
  \providecommand\BibTeX{{%
    \normalfont B\kern-0.5em{\scshape i\kern-0.25em b}\kern-0.8em\TeX}}}

\copyrightyear{2022}
\acmYear{2022}
\setcopyright{acmcopyright}\acmConference[SA '22 Conference Papers]{SIGGRAPH Asia 2022 Conference Papers}{December 6--9, 2022}{Daegu, Republic of Korea}
\acmBooktitle{SIGGRAPH Asia 2022 Conference Papers (SA '22 Conference Papers), December 6--9, 2022, Daegu, Republic of Korea}
\acmPrice{15.00}
\acmDOI{10.1145/3550469.3555389}
\acmISBN{978-1-4503-9470-3/22/12}

\def\authorsaddresses#1{\def\@authorsaddresses{#1}}
\authorsaddresses{\@mkauthorsaddresses}
\authorsaddresses{123456}

\begin{document}

\title{Shape Completion with Points in the Shadow}

\author{Bowen Zhang}
\affiliation{%
  \institution{Xi’an Jiaotong University}
  \city{Xi'an}
  \country{China}}
\email{zhangbowen.wx@gmail.com}

\author{Xi Zhao}
\authornote{Corresponding author: Xi Zhao, School of Computer Science and Technology, Xi'an Jiaotong University (zhaoxi.jade@gmail.com)}
\affiliation{%
  \institution{Xi’an Jiaotong University}
  \city{Xi'an}
  \country{China}}
\email{xi.zhao@mail.xjtu.edu.cn}

\author{He Wang}
\affiliation{%
  \institution{University of Leeds}
 \city{Leeds}
  \country{UK}
  }
\email{realcrane@gmail.com}

\author{Ruizhen Hu}
\affiliation{%
  \institution{Shenzhen University}
  \city{Shenzhen}
  \country{China}}
\email{ruizhen.hu@gmail.com}

\renewcommand{\shortauthors}{Zhang et al.}


\begin{abstract}
Single-view point cloud completion aims to recover the full geometry of an object based on only limited observation, which is extremely hard due to the data sparsity and occlusion. The core challenge is to generate plausible geometries to fill the unobserved part of the object based on a partial scan, which is under-constrained and suffers from a huge solution space. Inspired by the classic shadow volume technique in computer graphics, we propose a new method to reduce the solution space effectively. Our method considers the camera a light source that casts rays toward the object. Such light rays build a reasonably constrained but sufficiently expressive basis for completion. The completion process is then formulated as a point displacement optimization problem. Points are initialized at the partial scan and then moved to their goal locations with two types of movements for each point: directional movements along the light rays and constrained local movement for shape refinement. We design neural networks to predict the ideal point movements to get the completion results. We demonstrate that our method is accurate, robust, and generalizable through exhaustive evaluation and comparison. Moreover, it outperforms state-of-the-art methods qualitatively and quantitatively on MVP datasets.

\end{abstract}
\begin{teaserfigure}
\centering
\includegraphics[width=\linewidth]{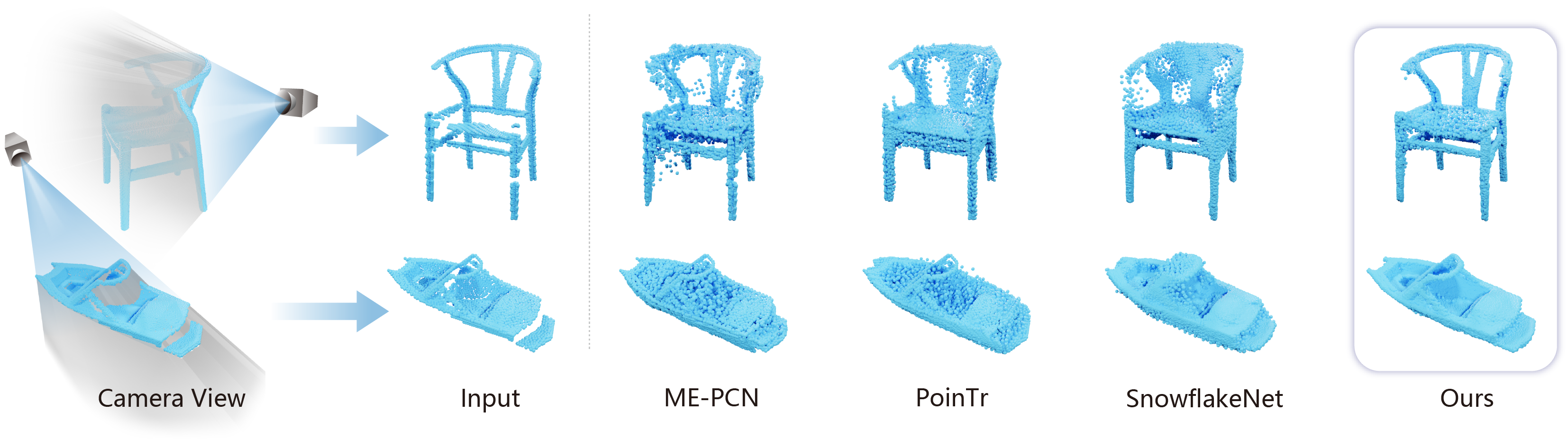}
\caption{Given a single-view partial scan, our method considers the volume that is shadowed by the observed points and generates more complete and clean results compared with ME-PCN~\cite{gong2021me}, PoinTr ~\cite{yu_pointr_2021} and SnowflakeNet~\cite{xiang_snowflakenet_2021}.}
\label{fig:teaser}
\end{teaserfigure}
\maketitle

\section{Introduction}

3D point cloud has become popular with the fast development of depth cameras and 3D scanning devices. It has been employed to acquire geometries from small objects ~\cite{zeisl_automatic_2013} to city-scale infrastructure ~\cite{lai_large-scale_2011} in various applications such as SLAM and self-driving cars. Although it is possible to acquire full observations via panoramic scanning in some scenarios ~\cite{kurkela_utilizing_2021}, very often, it requires a completion step as only partial scans can be obtained in many cases ~\cite{fei_comprehensive_2022}.

Point cloud completion aims to generate complete shapes from a partial point cloud. The partial point cloud can be any incomplete data, but a prevalent type is the partial scan, especially the single view scan. Very recently, 3D models have been used to synthesize single view partial scans for point cloud completion. The well-known completion benchmarks, such as ShapeNet ~\cite{chang_shapenet_2015}, Completion3D ~\cite{tchapmi_topnet_2019}, MVP ~\cite{pan_variational_2021}, et al. are all simulated from ShapeNet models and provide single-view partial scans and the ground-truth point cloud for machine learning. Based on these datasets, research on point cloud completion, especially deep learning-based techniques, has grown rapidly in recent years. However, existing methods almost entirely rely on network's capacity to learn the correlation between the observed partial scans and the complete ground truth. This leads to an under-constrained solution space for filling in the missing geometry. This uncontrolled data-fitting can generate out-of-distribution geometries far from the ground truth or average over multiple possible missing geometries related to the same input scan.

\begin{figure}[t]
\centering
\includegraphics[width=\linewidth]{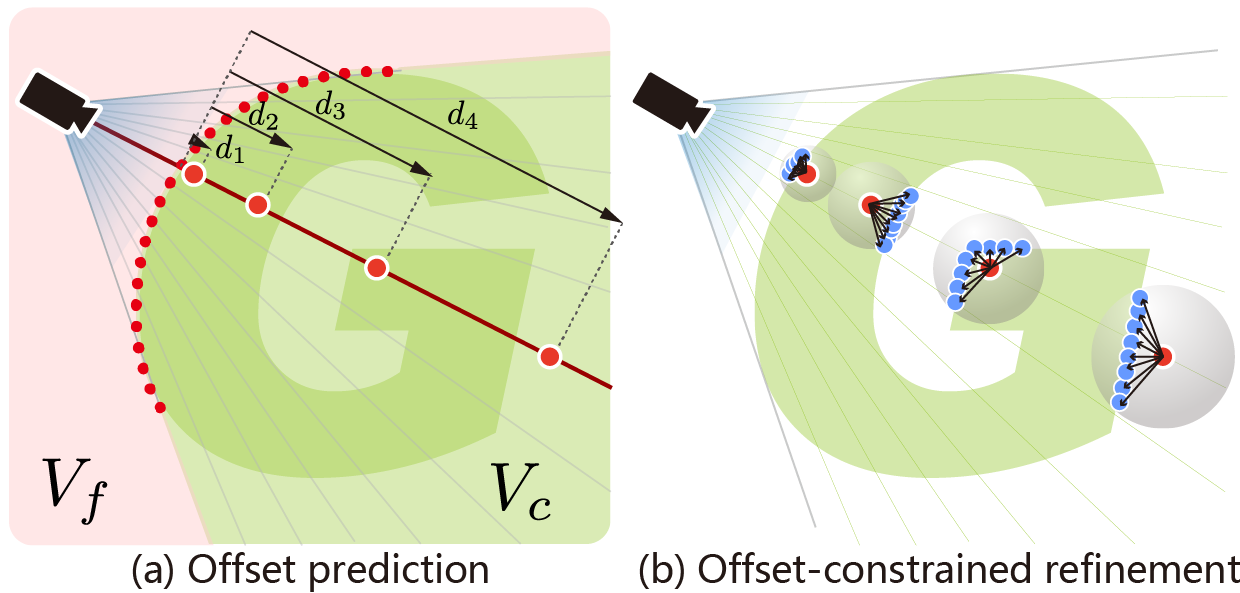}
\caption{
The completion along rays. (a)we duplicate the observed point multiple times to get the larger red points, and move them along the ray with offset $d_i$. (b) we split each red point multiple times to get the blue points and spread them to a neighborhood of the red point (the transparent circle).  
}
\label{fig:completion_process}
\end{figure}
Our insight is that the solution space should be restricted and more discriminative by leveraging prior knowledge. A key observation is that if we know the camera pose for the partial scan, the space for the completion can be reduced by explicitly modeling the spatial relation between the camera and the observed geometry. Inspired by a computer graphics technique called shadow volume ~\cite{crow_shadow_1977}, we can consider the camera as a point light facing the object, so that the unobserved geometry should be inside the volume that is shadowed by the object (the left column of Fig.\ref{fig:teaser}). This volume is naturally a reduced solution space for the missing geometry. To complete the geometry, we assume the observed part can shoot particles into the shadow volume, and they will eventually land on the missing surface of the object to help the shape completion. However, the shadow volume has an arbitrary shape, so we need to construct a convenient basis for possible particle movements.

We use a 2D example (Fig.\ref{fig:completion_process}) to demonstrate our main idea. We define the shadowed volume as ``candidate volume'' ($V_c$) for the completion (the green area in Fig.\ref{fig:completion_process}(a)). Inversely, the volume other than the candidate volume is the ``forbidden volume'' ($V_f$) for completion. 
Given a camera pose and the observed points $P$, We can approximate the candidate volume $V_c$ by casting rays from the camera to each observed point. These rays span the entire space of $V_c$. 
We consider each ray as a completion basis, and apply the completion process along each ray with the following two steps.
First, we put multiple duplicated points on each observed point and move them along the ray with distance $d_i$ to produce the initial guesses of the unobserved points (the red points). 
The second step is refinement. We split each red point into multiple blue points. The final locations of the blue points are computed by first duplicating the red points multiple times and then spreading them within the neighborhood of the red point. 
We do not simply move the blue points around locally, but constrain the moving range adaptively for each point. The points farther away from the observed points have larger moving ranges.

Traditionally, Chamfer Distance($CD$) is widely used for evaluating point cloud completion quality. However, $CD$ is a single value that is not sufficiently discriminative and fine-grained for detailed completion analysis. We argue that completion should be assessed on observed and unobserved parts separately because the two parts reflect different aspects of completion. The reconstruction of the observed part shows the \textit{fidelity} while the missing part shows the \textit{plausibility}. Therefore, in addition to the traditional metrics such as $CD$ and F-score, we employ a more fine-grained $CD$ metric that separately evaluates the reconstruction quality of the observed and unobserved parts.

Overall, our method can reduce the solution space with the guidance of the camera view. Furthermore, with the two types of movements inside the reduced solution space, our method can retain the observed geometric details and generate the missing part simultaneously.
Exhaustive evaluation and comparison show that our method outperforms the state-of-the-art methods both quantitatively and qualitatively. Formally, our contributions include:

1. A novel formulation of single-view point cloud completion with highly-reduced solution space;

2. An effective two-step completion method based on point movements with constrained direction and range;

3. An intuitive fine-grained metric that separately evaluates the observed and unobserved parts.

\section{Related work}
Among the huge number of point cloud completion methods, the methods based on the point cloud representation and deep learning network are most related to our work. Considering the completion methods from the input's point of view, we can roughly classify the existing methods into two categories: one is the methods that only use the input implicitly. This type of method produces typically the final results based on the guidance of a global feature. The other is the methods that retain the input explicitly for producing the results.
Besides, as we consider the completion process a displacement between point sets, we also introduce the displacement-related techniques for the point cloud.

\subsection{Global Feature Guided Completion}
This type of work extracts global features from partial input by using methods such as PointNet ~\cite{qi2017pointnet}, and then decodes this feature to generate the completion result. We call it ``implicit'' because the partial input affects the final results indirectly through the latent feature. 
Most point cloud completion work during the last few years falls to this type. The early methods such as ~\cite{yuan_pcn_2019} apply an encoder-decoder network. Tchapmi et al. ~\cite{tchapmi_topnet_2019} improved the decoder by using a hierarchical structure. Wen et al. ~\cite{wen_point_2020} further apply the skip-attention mechanism to convey geometric information from partial input to the hierarchical decoder to improve the results.
In ~\cite{xie_grnet_2020}, the authors use 3D grids as the intermediate representation to reduce the loss of structural and context details during completion.

These works use different ways to enforce the influence of structure and geometry details of partial input, but no matter what techniques they use, the influence of the guidance is limited. Because when generating complete shapes, the solution space is too large.

\begin{figure*}[thbp]
\centering
\includegraphics[width=\linewidth]{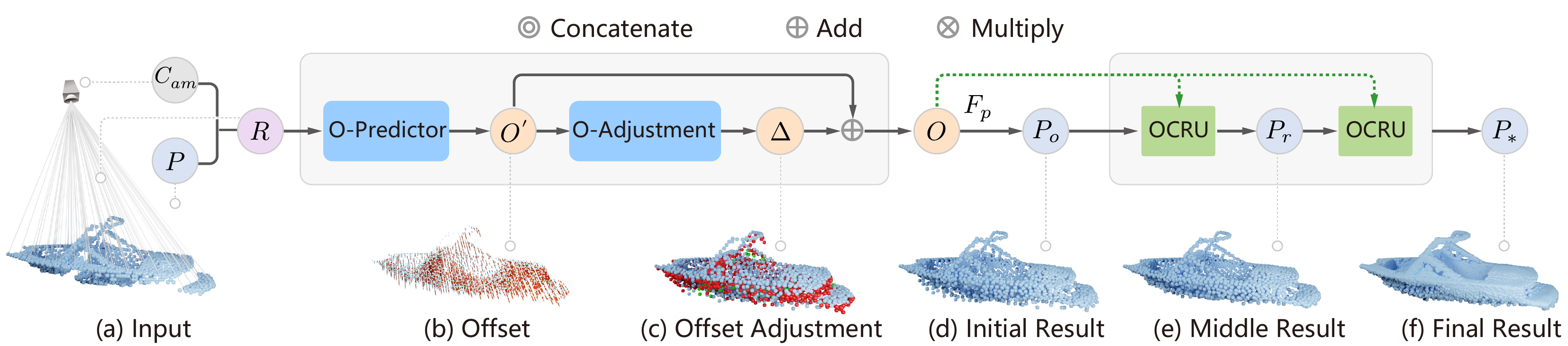}
\caption{The overview of our method. Given the camera location and partial scan, we compute a batch of rays that is the network's input(a). For each ray, we first predict offset $O'$ that is visualized by lines in (b), and then predict an adjustment for each offset that is visualized in (c), where green points represent positive adjustment; red points represent negative adjustment. With the initial completion result(d), we further apply a two-step refinement to get a smoother result (e) and final result (f).}
\label{fig:overview}
\end{figure*}
\subsection{Input Points Retained Completion}
To maximize the use of the input point cloud, researchers make efforts to retain the input points as much as possible. We classify the strategies of retaining input into two categories: keep all original input points in the result and merge the input into the middle result before refinement.

The most straightforward way to retain the input is to predict the missing part of the shape, and then put the input and generated part together to produce the completion result. The representative works of this kind are PF-Net ~\cite{huang_pf-net_2020} and PoinTr ~\cite{yu_pointr_2021}. PF-Net uses a multi-scale generating network to generate missing regions. PoinTr uses a transformer encoder-decoder architecture to predict the point proxies for the missing part. The benchmark they use normally contains a random removal of parts from the shape. Although completion for this type of partial data is useful, it is not as general as the completion of a partial scan point cloud.   

To deal with the single view partial scan data, Liu et al. ~\cite{liu_morphing_2019} propose to put the input partial scan and the initial completion results together to further feed to the refinement process. Normally there is also a sub-sampling step to control the number of points generated. Such a ``merge and sub-sampling''  strategy has become popular for strengthening the influence of input data on the results. It has been used in many works, such as ~\cite{zhang_detail_2020, wang_cascaded_2020, xia_asfm-net_2021, xiang_snowflakenet_2021, gong2021me}.  
The interesting extension for ``merge and sub-sampling'' is to use the symmetry axis of the shape, and apply the ``mirror'' operation before merging the input point cloud ~\cite {wang_cascaded_2020, xia_asfm-net_2021}.

Our method can be roughly considered as the second input retain strategy. However, we merge the input and the generated part differently. By predicting the offset of points along rays, both the observed and unobserved parts are processed uniformly in the system, so there is no need for the traditional ``merge and sub-sampling'' operation.

\subsection{Displacement-based Point Cloud Manipulation}

Yin et al. ~\cite{yin_p2p-net_2018} propose a one-step deformation to deform one point cloud to another by using a cycle structure network. This method can be generally applied to the transformation between two domains.
Another category of work is to use the point-wise displacement for manipulating the articulated point cloud shapes ~\cite{yan_rpm-net_2020, yi_deep_2018}. In ~\cite{yi_deep_2018}, the pair-wise displacement is used to find the correspondence between points from two sets and further predict the segmentation and part-based motion.
RPM-net ~\cite{yan_rpm-net_2020} predicts a temporal sequence of point-wise displacement for the input shape to infer movable parts and generate motions.

PMP-Net ~\cite{wen_pmp-net_2021},  PMP-Net++ ~\cite{pmpnet++} and Front2Back ~\cite{yao2020front2back} are more related to our work as they aims to do the completion. PMP-Net and PMP-Net++ enable multi-step movement of the input point set and use the least total moving distance loss to mimic the earth mover distance. 
Compared to them, we use viewpoint to reduce the solution space for the point’s displacement problem, and constrain the movement of points with the rays. Front2Back ~\cite{yao2020front2back} directly uses occlusion masks from the input and predicts the points on the ``other side''. It has difficulty in dealing with objects with complex structures as they cannot be simply described by opposite orthographic views. In contrast, our method is more flexible as we can generate different numbers of new points along the rays to better represent the geometry.

\section{Our Method}

\begin{figure*}[htbp]
\centering
\includegraphics[width=\linewidth]{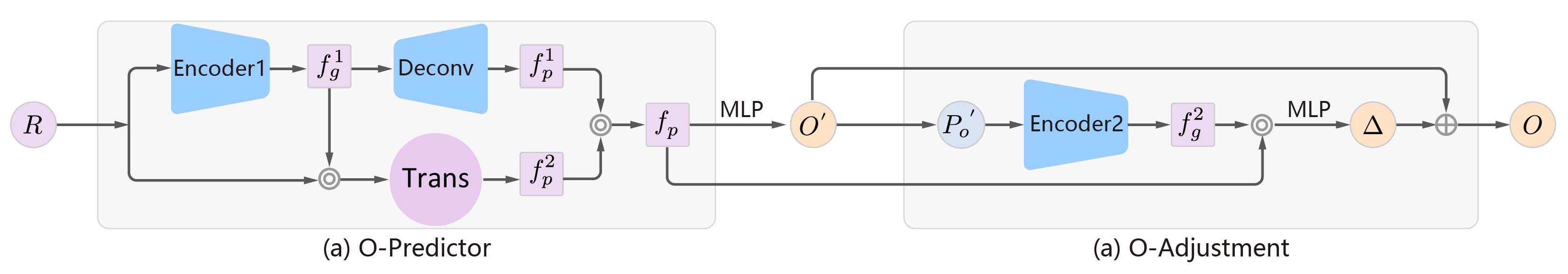}
\caption{The structure of the offset prediction module. It contains the initial offset prediction (O-Predictor)  and the offset adjustment (O-Adjustment).}
\label{fig:offset_prediction_and_adjustment}
\end{figure*}
\begin{figure}[tbp]
\centering
\includegraphics[width=\linewidth]{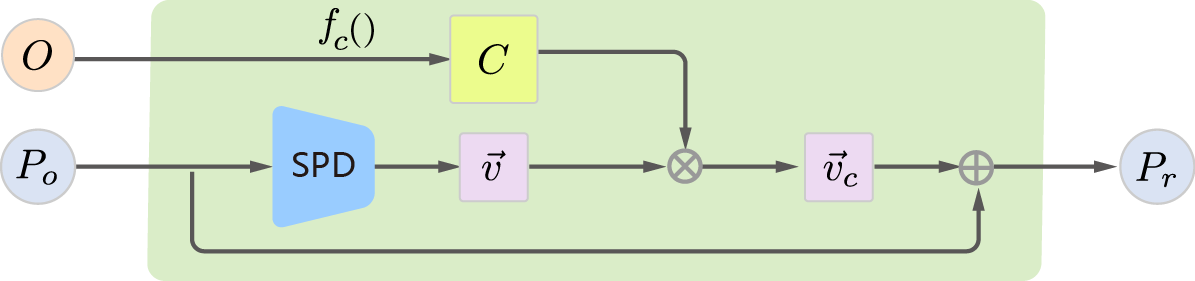}
\caption{The structure of offset constrained refinement unit (OCRU). We use function $f_c$ to compute the value $C$, which is used to constrain the range of movement $\vec{v}_c$ computed by SPD.}
\label{fig:offset_constrained_refinement}
\end{figure}
\subsection{Problem Formulation}

We formulate the completion process as a point displacement optimization problem. First, points are initialized at the partial scan. Then they are moved to their goal locations with two types of movements shown in Fig.\ref{fig:completion_process}. We design a network which consists of two main modules for such a completion process: the offset prediction module is designed to predict the movement along rays, then the offset-constrained refinement module is designed to estimate the local refinement movements. We show the overall architecture of our method in Fig.\ref{fig:overview}.

The input of our system contains a camera location, a partial scan and the rays. They are constructed as follows:
assuming the camera is located at a 3D position $Cam = (x_c, y_c, z_c)$, and faces the object center. From such a camera configuration, we can get a partial scan containing $N$ 3D points $P =\{p^i\}, (i = 1, ...,N)$. 
We define a set of rays along each $p^i$, which is represented by a vector from $Cam$ to $p^i$: $R= \{\vec{r_i}\} = \{p^i - Cam\}, (i = 1, ..., N)$.

\textbf{Offset prediction.} This module computes the initial completion result $P_o$ from partial scan $P$. We duplicate the input points multiple times and move these points along the rays with the predicted offset to get $P_o$. 
We can describe this process with:
\begin{equation}
\label{eq:initialDis}
    p_o^{i,l} = p^i + \mathcal{F}_{d}^{l}(p^i) \cdot \vec{r_i}
\end{equation}
where $i = 1,...,N; l = 1,...,L$. The function $\mathcal{F}_{d}:\mathbb{R}^3\rightarrow \mathbb{R}$ computes $L$ offsets for each camera ray. 
We provide more details of this module in Section 3.2.

\textbf{Offset-constrained refinement.} This module computes the final result $P_*$ from the initial completion result $P_o$. We split each point of $P_o$ into multiple points, and move them within a local neighborhood to improve the local geometry details. 
The equation of this step is:
\begin{equation}
    p_*^{m,k} = p_o^m + \mathcal{F}_{\vec{v}}^{k}(p_o^{m})
\end{equation}
where $m = 1, ..., N_o$, $k=1,...,L_*$. $N_o$ is the number of points in $P_o$. The function $\mathcal{F}_{\vec{v}} :\mathbb{R}^3\rightarrow \mathbb{R}^3$ is for computing $L_*$ local movements, in which the local movement of each point is constrained by its offset along the ray computed in the previous module.
We provide more details of this module in Section 3.3.

\subsection{Offset Prediction}

To estimate the offset mapping $\mathcal{F}_d$ defined in Equation~\ref{eq:initialDis}, we design a network which contains two parts: initial offset prediction (O-Predictor) and the offset adjustment (O-Adjustment). The network structure is shown in Fig.\ref{fig:offset_prediction_and_adjustment}. 

\textbf{O-Predictor.} 
Since $R$ contains not only the point positions, but also the orientations depending on the camera, modeling the correlations of both positions and orientations is crucial. We design a network module named O-Predictor to model the correlations for predicting offsets. The input of the network is ray $R$ and the output is $L$ offsets along each ray, denoted as $O' = \{d^{il}\} (i = 1, ... , N; l = 1, ... ,L)$, where each offset $d^{il}$ is a distance from $p^i$ to a new point along the $i_{th}$ ray in the positive direction.

Given rays $R$, we first extract the global correlations between all $\vec{r_i}\in R$ in a latent space using  PointNet++ ~\cite{qi_pointnet_2017} to get a global feature $f_g^1$, then transform the correlation into per-ray features $f_p^1$, both of which consider the orientation and position in $R$ simultaneously. In parallel, we concatenate $R$ and $f_g^1$ and pull it through a skip Transformer~\cite{xiang_snowflakenet_2021} where the attention is solely based on the orientation similarity between rays, to get the per-ray orientation correlation feature $f_p^2$. Finally, $f_p^1$ and $f_p^2$ are concatenated with rays $R$, partial scan $P$ and global feature $f_g^1$ to compute a feature $f_p$ which is further fed into a MLP to compute the output offset $O'$. We apply Relu to the output of MLP to make sure all the offset values are non-negative.

\textbf{O-Adjustment.}
O-Predictor alone can only provide rough offset estimation because learning the complex distribution of offsets for different shapes is non-trivial. We design an O-Adjustment module to improve the offset precision by modifying the predicted offset values. From the example shown in  Fig.\ref{fig:overview}, we can see that although the points at the top of the watercraft move too much away from the bar after the O-Predictor step(b), the O-Adjustment module move them back by assigning these points with negative adjustment(c).

The structure of the O-Adjustment module is as follows. After getting the offsets $O'$ from the last module, we compute a first step completion results $P_{o'}$ by moving each input point with this offset along its ray. This process can be presented as 
$ p^i + d^{il} \cdot \vec{r_i}$.
Then we use the PCN encoder ~\cite{yuan_pcn_2019} to extract the global feature $f_g^2$, which is concatenated with feature $f_p$ and fed to MLP to predict an adjustment for each offset, denoted as $\Delta$. The final offset $O$ is the sum of ${O'}$ and $\Delta$ . We again apply the Relu function to $O$ to make sure the output offset is non-negative values. Finally, we convert the output of the offset $O$ to the initial completion result $P_o$ by moving points along rays.

\subsection{Offset Constrained Refinement}

The main problems with the initial result $P_o$ are the non-uniform density and lack of local geometry details. The points in the input partial scan $P$ may distribute unevenly due to the camera configuration relative to the shape, which results in non-uniformed distribution of corresponding rays. In addition, the points farther away from the camera are sparser because of the perspectivity of the rays. Therefore, refinement is needed to generate fine-grained geometry structures.

Preliminary experiments show that classic refinement methods such as folding operation ~\cite{yang_foldingnet_2018} and the snowflake point deconvolution (SPD)~\cite{xiang_snowflakenet_2021} tend to fill holes. However, if the holes are part of the object, filling them can damage the topology and structure of the shape. 
Besides, their refinement process cannot control the movement ranges of the points. For example, a point may move to the invalid space of the partial scan $P$, which leads to a noisy effect. 

To address this problem, we propose an offset-constrained refinement module (Fig.\ref{fig:offset_constrained_refinement}) to constrain the movement range during the refinement.
We first apply a farthest point sampling (PFS) for $P_o$ to make the points distribute more evenly. Then the result is further fed to the offset constrained refinement module, which contains two layers of refinement. Each layer is an offset-constrained refinement unit (OCRU) designed based on SPD. Here the output of SPD is a list of 3D movement vectors. We multiply the offset constraint value to each dimension of the SPD's output to apply the constraint. 
More specifically, the offset constraint value for $j_{th}$ point in $u_{th} (u = 1,2)$ OCRU is computed by:
\begin{equation}
C_u^j = f_c(O_{j}, u) = (O_{j}/2 + 0.03)/\alpha^{u-1}
\end{equation}
where $O_j$ is the overall offset for $j_{th}$ point computed by the offset prediction module. $\alpha$ is a scale coefficient. In all the experiments, we set $\alpha = 1.5$.

With this constraint, the local details of the partial scan are kept. In the final result $P_*$, the points close to partial scan $P$ have tiny offsets (the offsets are nearly zero), so the movement range of such points in the refinement module is very small. As a result, the observed geometry represented by such points is retained. 
The boundary points of the partial scan are also part of the observed points, so the generated points stem from them should not move far too. As a result,  the geometry in the boundary area is also neat and clean. The movement range is more extensive for points generated around the far end of the rays. So points have more freedom to move during refinement. With such a strategy, we effectively retain the observed geometry and improve the unobserved part's quality.

\subsection{Loss Function}
We apply the shape loss $\mathcal{L}_{CD}$, which is Chamfer Distance ($CD$) to measure the shape difference between the completed results and the ground truth. The $CD$ between point cloud $P_1$ and $P_{2}$ is computed by:
\begin{equation}
     \mathcal{L}_{CD}(P_1, P_2) = \sum_{p \in P_1} \min_{q \in P_2} ||p-q||^2_2 + \sum_{q \in P_2} \min_{p \in P_1} ||q-p||^2_2
\end{equation}

During the training, we first pre-train the offset prediction module with loss function $\mathcal{L}_{CD}(P_{o'}, P_{\text{gt}_1}) + \mathcal{L}_{CD}(P_{o}, P_{\text{gt}_1})$, where the $P_{o'}$ and $P_{o}$ are computed by the offset prediction module. Then, we keep the parameters in the offset prediction module fixed and pre-train the refinement module with loss function 
$\mathcal{L}_{CD}(P_{r}, P_{\text{gt}_2}) + \mathcal{L}_{CD}(P_{*}, P_{\text{gt}_3})$, where $P_{r}$ and $P_{*}$ are the middle and final result of the refinement module. Finally, we train two modules jointly with the same loss function as the second stage. $P_{\text{gt}_1}$, $P_{\text{gt}_2}$ and $P_{\text{gt}_3}$ are the ground truth point cloud which contains 8192, 2048 and 16384 points respectively. 
We provide other training details in the supplementary material.

\begin{figure}[thbp]
\centering
\includegraphics[width=0.99\linewidth]{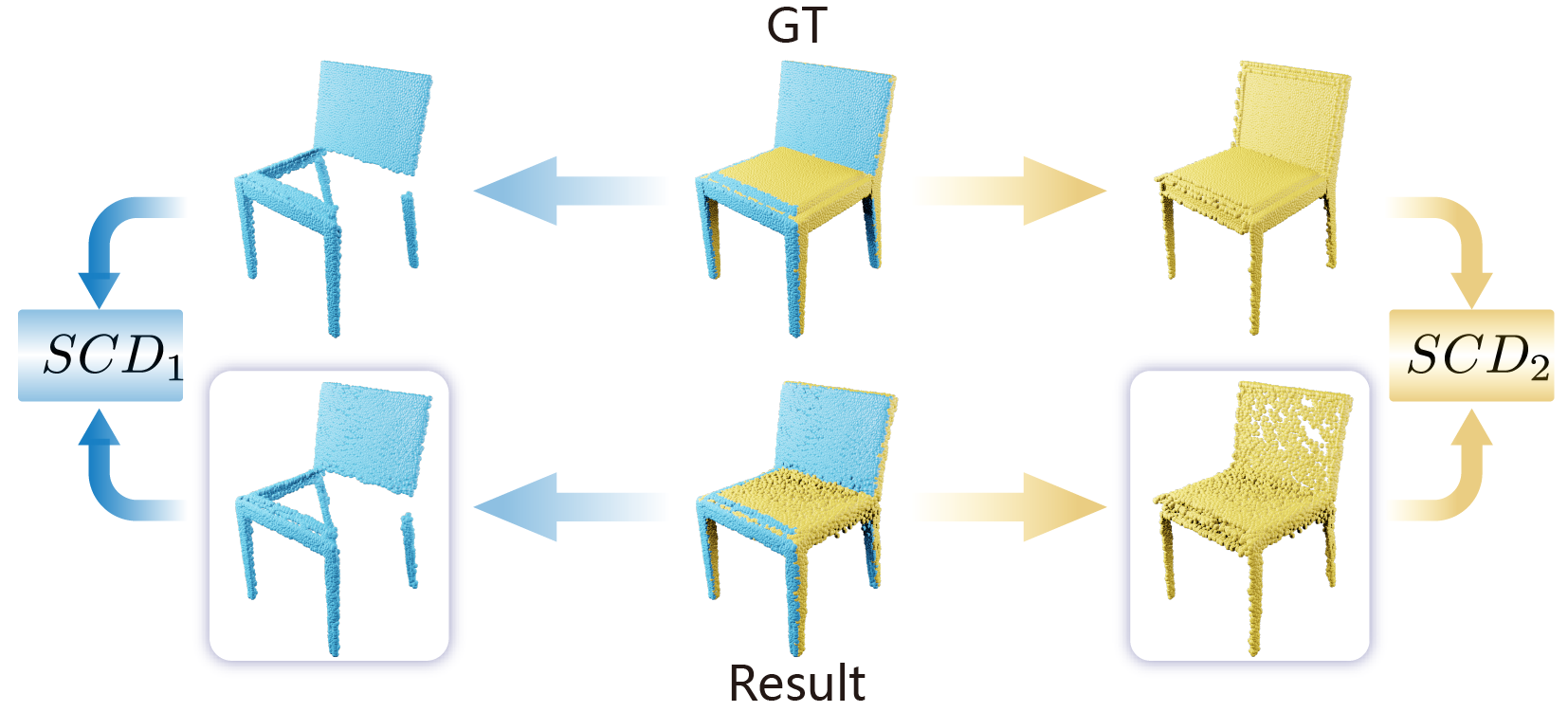}
\caption{The computation of $SCD_1$ and $SCD_2$.}
\label{fig:Computation_of_SCD}
\end{figure}
\begin{figure*}[htbp]
\centering
\includegraphics[width=\linewidth]{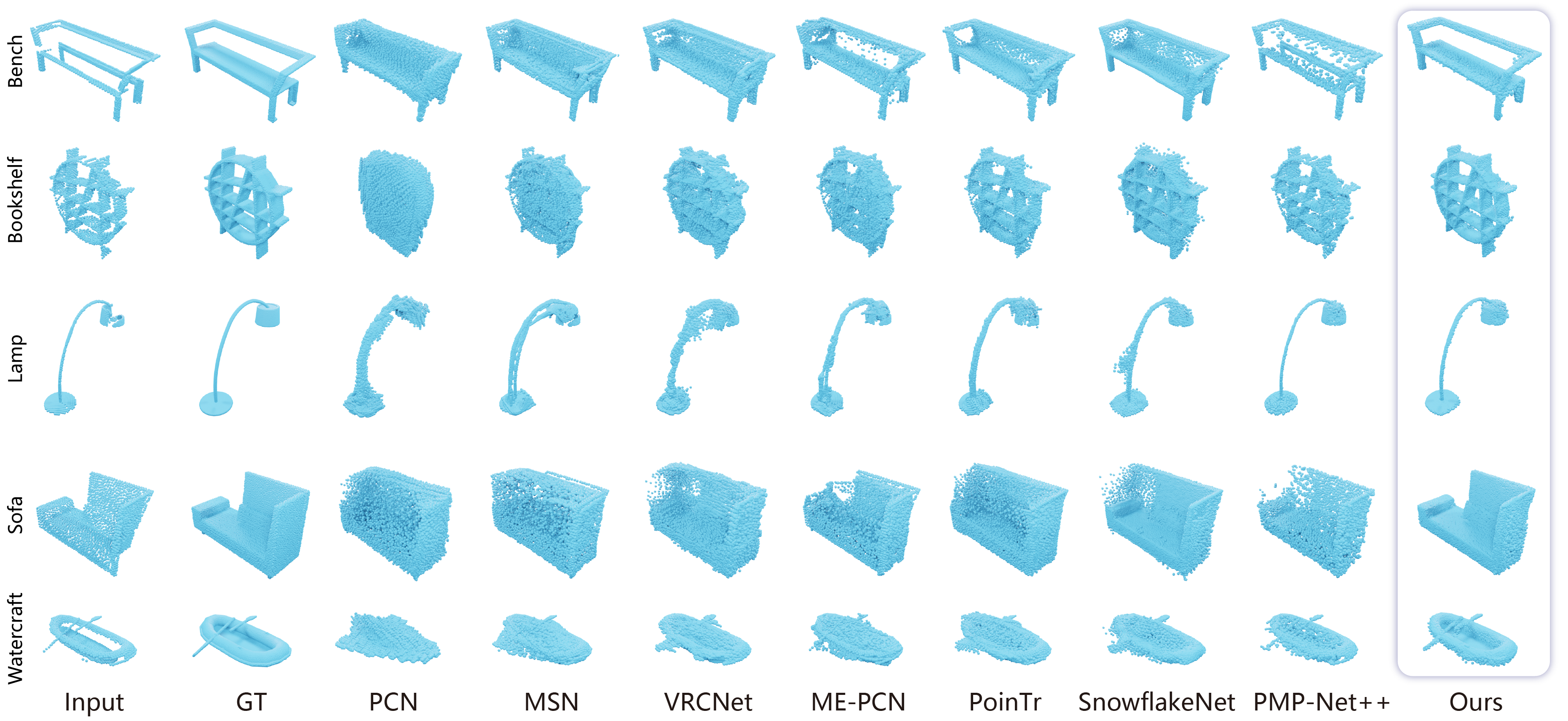}
\caption{Visualized completion results for comparison on MVP dataset.}  
\label{fig:overall_comparison}
\end{figure*}

\section{Experiments}
\subsection{Dataset}
We use the Multi-View Partial point cloud dataset (MVP) ~\cite{pan_variational_2021} in our experiments. MVP is computed based on 16 categories of 4000 CAD models. Twenty-six camera locations are sampled for each model to simulate the partial scan. In our experiment, we recompute the MVP dataset to record the camera configuration for each scan. Finally, we use 62400 partial-complete point cloud pairs for training and 41600 pairs for testing. 

\subsection{Evaluation Metric}

We use $CD$, F-Score ~\cite{tatarchenko_what_2019} and Density-aware $CD$ ($DCD$) ~\cite{wu_density-aware_2021} to evaluate the completion results.
$CD$ is sensitive to outliers but insensitive to the local density. F-score is used as a supplement to $CD$ to evaluate the completion results. A higher F-score usually indicates better visual quality. $DCD$ is also a metric for evaluating the overall point cloud distance, which is more sensitive to the density distribution of point clouds.

The goal of the completion task naturally includes two parts: retaining the geometry of the partial scan (task1) and generating a reasonable shape for the unobserved part (task2). 
Existing metrics are indiscriminative to the two parts, which suggests that they lack granularity.
To provide sufficient details on the completion quality, we propose a new metric named ``split $CD$'' ($SCD$) that includes two values, $SCD_1$ and $SCD_2$. 

To compute $SCD_1$ and $SCD_2$, we split the GT point cloud $P_\text{gt}$ into two parts $P_\text{gt}^1$ and $P_\text{gt}^2$ based on the partial input $P$. More specifically, if a point in $P_\text{gt}$ is in the neighbourhood of $P$, then it belongs to $P_\text{gt}^1$, otherwise it belongs to $P_\text{gt}^2$. For each point $p$ in $P$, the neighborhood of $p$ is defined as the spatial area within radius $r$ ($r=0.01$ in our setting), and the neighborhood of $P$ is the union of the neighborhoods of all $p$ in $P$.
We next split the completion results into two parts based on $P$ and $P_\text{gt}^2$: for each point $p$ in a result $P_*$, we find a closest point $p'$ in the point set $\{P, P_\text{gt}^2\}$. If $p'$ belongs to $P$, then $p$ belongs to $P_{*}^1$, otherwise $P_{*}^2$. Then the split $CD$ is computed as 
$SCD_1 = \mathcal{L}_{CD}(P_*^1, P_\text{gt}^1)$, 
$SCD_2 = \mathcal{L}_{CD}(P_*^2, P_\text{gt}^2)$.
This process is visualized in Fig.\ref{fig:Computation_of_SCD}.

\subsection{Comparison}
We exhaustively compare our method with seven baseline methods: PCN~\cite{yuan_pcn_2019}, MSN~\cite{liu_morphing_2019}, ME-PCN~\cite{gong2021me}, VRCNet~\cite{pan_variational_2021}, PoinTr~\cite{yu_pointr_2021},
SnowflakeNet~\cite{xiang_snowflakenet_2021}, 
and PMPNet++~\cite{pmpnet++}. 

\setlength{\tabcolsep}{4pt}
\begin{table}
\begin{center}
\caption{Evaluation for each method. $CD$ and $SCD$ are multiplied by $10^4$. 
}
\label{table:quantitative_comparison}
\begin{tabular}{l c c c c c}
\hline\noalign{\smallskip}
 Methods & $CD$ & F-Score & $DCD$ & $SCD_{1}$ & $SCD_{2}$\\
\noalign{\smallskip}
\hline
\noalign{\smallskip}
PCN         & 5.907	& 0.617	& 0.628	& 5.863	& 6.322 \\
MSN        & 4.749	& 0.682	& 0.645	& 2.210 & 7.757	\\
ME-PCN     & 4.680	& 0.662	& 0.658	& 2.086 & 7.341	\\
VRCNet      & 4.780 & 0.741 & 0.539 & 2.471 & 6.350 \\
PoinTr     & 3.882 & 0.715 & 0.613 & 1.709 & 6.664 \\
PMP-Net++   & 3.381 & 0.687 & 0.696 & 1.391 & 6.954 \\
SnowflakeNet & 2.696	& 0.796	& 0.524	& 0.951 & 4.683 \\
\hline
Ours   & \bf{2.419}    & \bf{0.800} & \bf{0.513} & \bf{0.646} & \bf{3.896} \\
\hline
\end{tabular}
\end{center}
\end{table}
\setlength{\tabcolsep}{1.4pt}
\setlength{\tabcolsep}{4pt}
\begin{table}
\begin{center}
\caption{Ablation study. $CD$ and $SCD$ are multiplied by $10^4$.
}
\label{table:ablation_all}
\begin{tabular}{l c c c c c}
\hline\noalign{\smallskip}
 Methods & $CD$ & F-Score & $DCD$ & $SCD_{1}$ & $SCD_{2}$\\
\noalign{\smallskip}
\hline
\noalign{\smallskip}
No GT viewpoint      & 2.635	& 0.781	& 0.538	& 0.678	& 4.590 \\
No adjustment    & 2.523	& 0.787	& 0.532	& 0.706 & 4.090	\\
No offset constraint    & 2.730	& 0.779	& 0.540	& 0.827 & 4.355 \\

\hline
Our final                   & \bf{2.419}	& \bf{0.800} & \bf{0.513} & \bf{0.646} & \bf{3.896} \\
\hline
\end{tabular}
\end{center}
\end{table}
\setlength{\tabcolsep}{1.4pt}
\setlength{\tabcolsep}{4pt}
\begin{table}
\begin{center}
\caption{Ablation study on the number of points along rays in the offset prediction module. $CD$ and $SCD$ are multiplied by $10^4$.}
\label{table:initial_offset_num_ablation}
\begin{tabular}{l c c c c c}
\hline\noalign{\smallskip}
Point number & $CD$ & F-Score & $DCD$ & $SCD_{1}$ & $SCD_{2}$\\
\noalign{\smallskip}
\hline
\noalign{\smallskip}
2   & \bf{2.375}	& 0.798 & 0.523 & \bf{0.608} & 4.066 \\
4 (Ours)   & 2.419	& \bf{0.800} & \bf{0.513} & 0.646 & \bf{3.896} \\
6   & 2.480	& 0.786	& 0.535	& 0.741 & 3.956	\\
\hline
\end{tabular}
\end{center}
\end{table}
\setlength{\tabcolsep}{1.4pt}

\textbf{Quantitative comparison.}
The quantitative results are reported in Table~\ref{table:quantitative_comparison}.
This table lists averaged metric values for all categories of shapes, and more detailed results are in the supplementary material. Overall, our method achieves the best performance in all metrics across all categories of objects. Lower $CD$ and higher F-Score values indicate an overall better completion quality. A lower $DCD$ value demonstrates that our method does not suffer from unbalanced local density compared with other baseline methods. 
The $SCD$ show a more fine-grained analysis with $SCD_1$ and $SCD_2$. PCN has similar scores for the two parts, which suggests that the completion quality for the observed part and the unobserved part are quite similar. For other baseline methods, the $SCD_1$ values are reduced, but the $SCD_2$ values do not change much (sometimes even increase). This reveals that these methods implicitly focus on retaining the observed geometry, sometimes at the cost of the completion quality of the unobserved part. One possible reason is that they allocate too many points to input to ensure reconstruction but fewer points for the missing part. 
Finally, SnowflakeNet does reduce both of these values but not significantly on $SCD_2$ compared with our method. Nevertheless, our method reduces both $SCD$ values, demonstrating that our results are not only overall better, but also have balanced improvement on both the observed and unobserved geometries.

\textbf{Qualitative comparison.}
In Fig.\ref{fig:overall_comparison}, we visually compare our method with other baseline methods on the MVP dataset. In general, our method can keep the geometric details of the partial input better. First, the partial scan boundary is not blurred during completion, which is crucial for model details such as the holes in the back of the chair, the bookshelve and the neck of the desk lamp. Besides, the concave areas in the input are also kept, such as the sitting area of the watercraft. The holes, boundaries and concave areas are vital geometric/topological features of an object, indicating their functionality or distinguishing them from similar objects. Keeping these details intact in the input is crucial for high-fidelity completion. Next, Our method can capture overall shape variations well across various shapes. It can deal with unique shapes that are rare in the training data. One example is the sofa with a rather asymmetric design, which is statistically rare in the data. However, our method can recognize and capture the overall asymmetry and recover the model precisely. In contrast, all other methods try to make the sofa symmetric during completion. 

We show results with panoramic views in the supplementary video and more completion results for all categories in the supplementary material.

\subsection{Ablation Studies}

Here we evaluate the design of the network of our method. We consider the following ablation versions:
\begin{itemize}
    \item  \textbf{No GT viewpoint.} The main assumption of our method is that the viewpoint configuration is known. We assume that we do not have this information and predict the viewpoint from the partial scan. 
    \item \textbf{No offset adjustment.} We test how our method performs when the offset adjustment step is removed.
    \item  \textbf{No offset constraint.} We test how our method performs if we do the refinement without the offset constrain.
    
\end{itemize}

The quantitative results are shown in Table~\ref{table:ablation_all}.
We can see that the final version of the method has the overall best performance. 

Without the viewpoint information, the quality of the final results is getting slightly worse. However, the average $CD$ and $SCD$ values are still better than all other baseline methods. Note that the prediction of viewpoint from partial input is not the focus of this paper. We found that with the imperfect viewpoint, our method still works and can produce results with satisfactory quality. The detail of the viewpoint prediction method and related experimental results can be found in the supplementary material.

\begin{figure}[tbp]
\centering
\includegraphics[width=\linewidth]{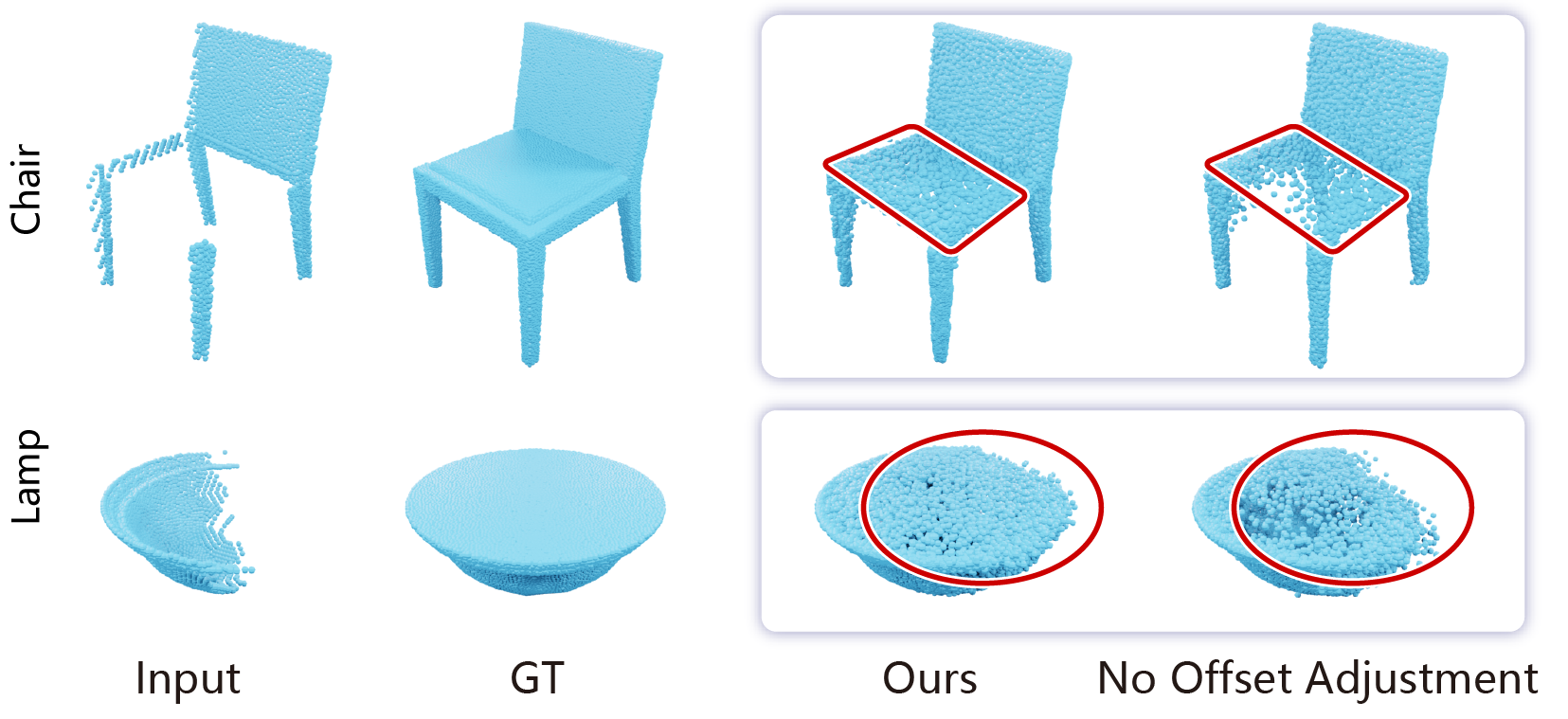}
\caption{Ablation study: compare our method with the version that has no offset adjustment step.}
\label{fig:no_offset_adjustment_comparison}
\end{figure}
When there is no offset adjustment step, the performance of our method drops. Some points cannot move to expected locations without the offset adjustment with a single step. Then the network tends to assign a small offset to points, so fewer points are generated in the missing area. This is why the $SCD_2$ is getting worse. In Fig.\ref{fig:no_offset_adjustment_comparison} we can see that the one-step offset results look crude: there are holes (such as the chair) or incomplete parts (such as the half-size lamp) in the unobserved area. With the offset adjustment step, the completion results $P_{o'}$ and $P_o$ are improved simultaneously, and so are the final completion results.

\begin{figure}[tbp]
\centering
\includegraphics[width=\linewidth]{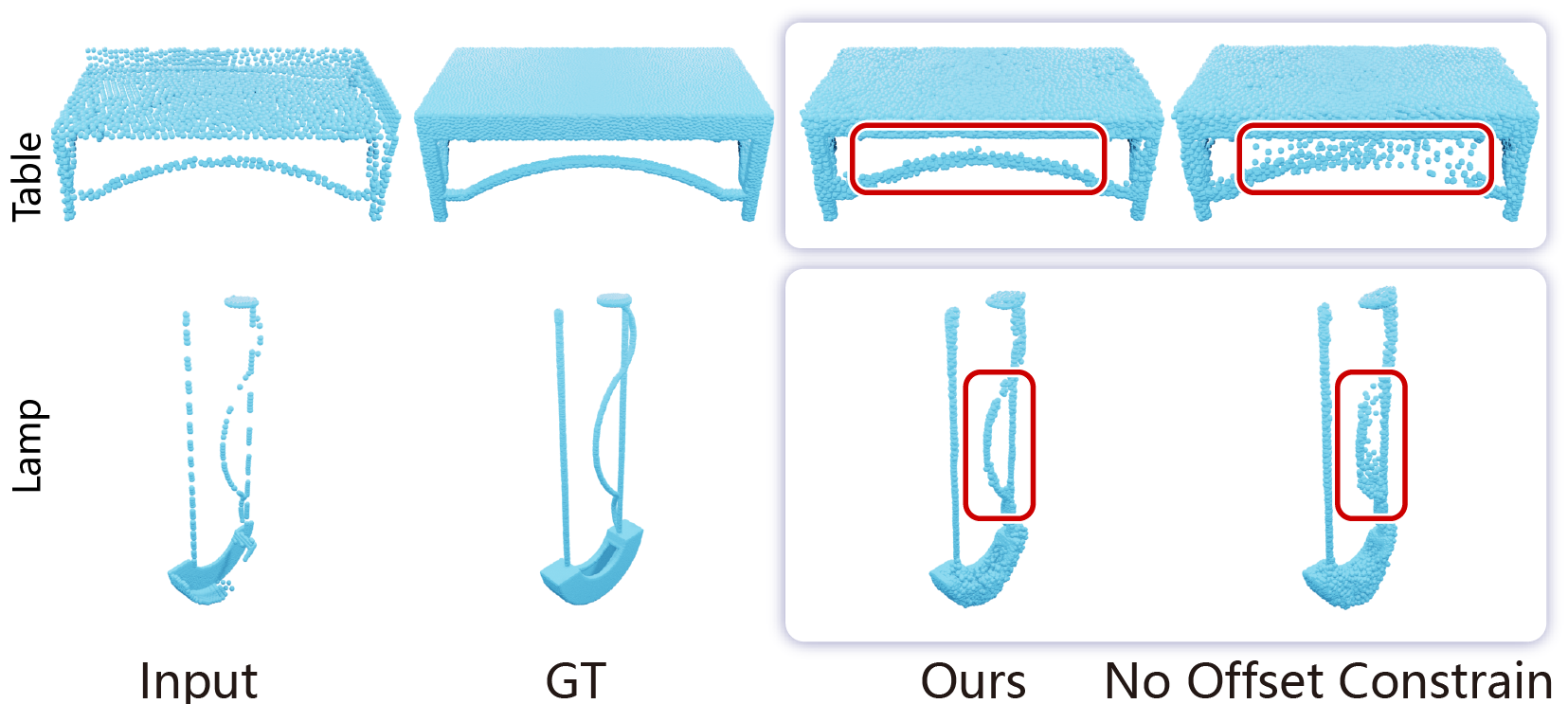}
\caption{Ablation study: compare our method with the version that does not apply offset constrain during refinement.}
\label{fig:no_offset_constrain_comparison}
\end{figure}
When there is no offset constraint in the refinement module, the performance of our method also drops. We can see that the $SCD_1$ value increases the most compared with other alternative solutions. Without the constraints, the points can move in larger ranges during refinement. Such movement especially makes the completion of the observed part worse because the local details tend to be blurred. We can see such effects in Fig.\ref{fig:no_offset_constrain_comparison}. Without the constraints, the area under the table surface and the details of the lamp are blurred, while with the constraints, we get very neat and clean local details. 

The number of points that move along each ray is also important for our method. We evaluate how many points should move along each ray during the completion. Theoretically, too few points  may not be enough to represent an initial complete result, because the rays can be sparse. However, too many points may cause unnecessary fillings inside the object, leading to larger shape loss and higher computation complexity. So we try different numbers of points and compute the evaluation values for each case. The results are shown in Table~\ref{table:initial_offset_num_ablation}. We found that using four points gets the overall best results with higher $DCD$, $SCD_2$ and F-score. Therefore, we use four points when implementing our experiments.

We also test our method with ablations in the offset prediction and adjustment steps regarding the transformer and the feature extraction. 
The details of these experiments can be found in the supplementary material.

\section{Conclusion}
This paper proposes a new framework for single view point cloud completion. The key insight is that the vast solution space can be drastically reduced when using shadow volumes and camera rays as the solution basis. This construction leads to two types of completion strategies by moving points for completion: one is moving points along rays, and the other is the local constrained movement for refinement. Moreover, we propose a more fine-grained metric for evaluation: the split $CD$ ($SCD$), with which we can analyze the completion quality for the observed part and the unobserved part separately and jointly. Extensive evaluation and comparison demonstrate the superiority of our method over the current SOTA methods.

\textbf{Limitations and future work.} Our method relies on the details of the partial scan. If the quality of the scan is low, especially when the main structure or significant part of the shape is not captured, our method might have difficulties recovering the correct local details. Two representative examples are shown and discussed in the supplementary material. In the future, instead of generating one geometry, we will aim to generate a distribution of possible geometries where more prior knowledge or user’s preferences can be considered. As mesh is a more useful data type, predicting implicit function along camera rays for reconstruction is also worth further exploration.

\begin{acks}
This work was supported in part by 
National Natural Science Foundation of China (62072366, 61872250), 
Key R\&D project of Shaanxi Province (2021QFY01-03HZ),
China Postdoctoral Science Foundation Funded Project (2020M673407), 
Guangdong Natural Science Foundation (2021B1515020085) and 
Shenzhen Science and Technology Program (RCYX20210609103121030).
\end{acks}

\bibliographystyle{ACM-Reference-Format}
\bibliography{egbib}

\end{document}